\documentclass[11pt]{article}
\usepackage{geometry}
\usepackage{coling2020}
\usepackage{times}
\usepackage{url}
\usepackage{latexsym}
\usepackage{microtype}
\usepackage{graphicx}
\usepackage{tabularx}
\usepackage{booktabs}
\usepackage{multirow}
\usepackage{algorithm} 
\usepackage{algpseudocode} 
\usepackage{amssymb,amsmath}
\usepackage[left]{lineno}
\usepackage{xcolor}
\colingfinalcopy 

\title{BPGC at SemEval-2020 Task 11: Propaganda Detection in News Articles with Multi-Granularity Knowledge Sharing and Linguistic Features based Ensemble Learning}

\author{Rajaswa Patil\textsuperscript{1} \and Somesh Singh\textsuperscript{2} \and Swati Agarwal\textsuperscript{2} \\
 \textsuperscript{1}Department of Electrical \& Electronics Engineering \\
 \textsuperscript{2}Department of Computer Science \& Information Systems \\
 BITS Pilani K. K. Birla Goa Campus, India \\
 {\tt \{f20170334, f20180175, swatia\}@goa.bits-pilani.ac.in}}
 
\date{}

\begin{document}
\maketitle
\begin{abstract}
Propaganda spreads the ideology and beliefs of like-minded people, brainwashing their audiences, and sometimes leading to violence. SemEval $2020$ Task-11 aims to design automated systems for news propaganda detection. Task-11 consists of two sub-tasks, namely, Span Identification - given any news article, the system tags those specific fragments which contain at least one propaganda technique and Technique Classification - correctly classify a given propagandist statement amongst $14$ propaganda techniques. For sub-task $1$, we use contextual embeddings extracted from pre-trained transformer models to represent the text data at various granularities and propose a multi-granularity knowledge sharing approach. For sub-task $2$, we use an ensemble of BERT and logistic regression classifiers with linguistic features. Our results reveal that the linguistic features are the reliable indicators for covering minority classes in a highly imbalanced dataset.
 
\end{abstract}

\section{Introduction}\label{sec:introduction}
\blfootnote{
    \hspace{-0.65cm}  
    This work is licensed under a Creative Commons 
    Attribution 4.0 International License.
    License details:
    \url{http://creativecommons.org/licenses/by/4.0/}.
}

Propaganda is biased information that deliberately propagates a particular ideology or political orientation \cite{aggarwal2019nsit}. Propaganda aims to influence the public's mentality and emotions, targeting their reciprocation due to their personal beliefs \cite{jowett2018propaganda}. News propaganda is a sub-type of propaganda that manipulates lies, semi-truths, and rumors in the disguise of credible news \cite{doi:10.1177/0896920518764586}. The emphasis on this manipulation differentiates propaganda and its various classes from each other and free verbalization \cite{hackett1984decline}. News propaganda can lead to the mass circulation of misleading information, shared agenda, conflicts, religious or ethnic reasons, and can further even lead to violence and terrorism. Due to massive size, high velocity, rich online user interaction, and diversity, the manual identification of propaganda techniques is overwhelmingly impractical. Hence, the development of a generalized system for propaganda detection in news articles is a vital task for security analysts and society \cite{barron2019proppy}. 

In \textit{SemEval2020-Task11}, \newcite{DaSanMartinoSemeval20task11} propose a corpus of $550$ news articles for propaganda detection. Each article is annotated with propaganda spans belonging to $14$ propaganda techniques. The annotation is performed at the fragment level. The task of propaganda detection is divided into two sub-tasks; Span identification (sequence labeling) and technique classification (sequence classification). Span identification sub-task aims to detect the propagandist spans of text in the news articles. Whereas, the technique classification sub-task aims to classify the propaganda spans into various propaganda techniques. The work presented in this paper aims to provide independent approaches for both sub-tasks $1$ and $2$. Most recent approaches for propaganda detection task use pre-trained transformer models. This work propose a multi-granularity knowledge sharing model built on top of the embeddings extracted from these transformer models for the propaganda span detection sub-task. Further, we show the effectiveness of ensembling linguistic features-based machine learning classifiers with these transformer models covering the minority-classes for the technique classification sub-task.

\section{Background}\label{sec:background}
Before the ascension of the detection of fake news and propaganda in NLP, \newcite{mihalcea2009lie} introduced the automatic detection of lying in text. In this work, the authors proposed three datasets based on \textit{abortion, death penalty, and best friends} and approached the multi-class classification problem. The Amazon Turk Service performed the annotations for these datasets. They used support vector machine and naive bayes classifiers for conducting their experiments. On the other hand, \newcite{ciampaglia2015computational} had employed knowledge graphs for computational fact-checking and utilizing data amassed from Wikipedia. Detection of propaganda in news articles was advocated by \newcite{rashkin2017truth} and \newcite{barron2019proppy}. The former proposed the use of Long Short-term Memory (LSTM) models and machine learning methods for deception detection and classification to different types of news, \textit{trusted, satire, hoax, and propaganda}. The latter presented \textit{proppy}\footnote{Proppy, propaganda detection system. \url{https://proppy.qcri.org/}} the first publicly available real-world and real-time propaganda detection system for online news. Fake news and propaganda have been getting more attention recently \cite{bourgonje2017clickbait,helmstetter2018weakly,jain2018fake}. Previous research has leveraged BERT \cite{devlin2018bert} to extract features for fake news detection \cite{8919302}. The constrained resources and lack of annotated corpora are considered to be an immense challenge for researchers in this field. \newcite{da-san-martino-etal-2019-fine} proposes a new dataset with text articles annotated on the fragment level with one of $18$ given propaganda techniques. To classify them, they also employ BERT models for this high granularity task. Given the relatively low count of some of these $18$ techniques, they merged similar underrepresented techniques into superclasses like \textit{Bandwagon, Reductio\_ad\_hitlerum}, \textit{Exaggeration,\ Minimisation}, \textit{Name\_Calling,\ labelling}, and \textit{Whataboutism,\ Straw\_Men,\ Red\_Herring} and eliminate \textit{Obfuscation, Intentional Vagueness, Confusion} to compile $14$ classes \cite{DaSanMartinoSemeval20task11}. 

\section{Methodology}\label{sec:methodology}
In this section, we explain our approach and the road-map to our conclusion. A detailed explanation of our implementations is provided in Section \ref{sec:experimental_setup}. Inspired by \newcite{da-san-martino-etal-2019-fine}, we consider three granularities of propaganda detection: Document-level, Sentence/Fragment-level, and Token-level. The Span Identification sub-task focuses on high-granularity propaganda detection through a token-level sequence labeling task. Whereas, the Technique Classification sub-task focuses on sentence/fragment level classification. Even though both the sub-tasks are restricted to a single granularity, the context from other granularities can be used by defining auxiliary objectives for each granularity.

\subsection{Sub-Task 1: Span Identification}
\begin{figure}[htbp]
    \center
    \includegraphics[scale=0.95]{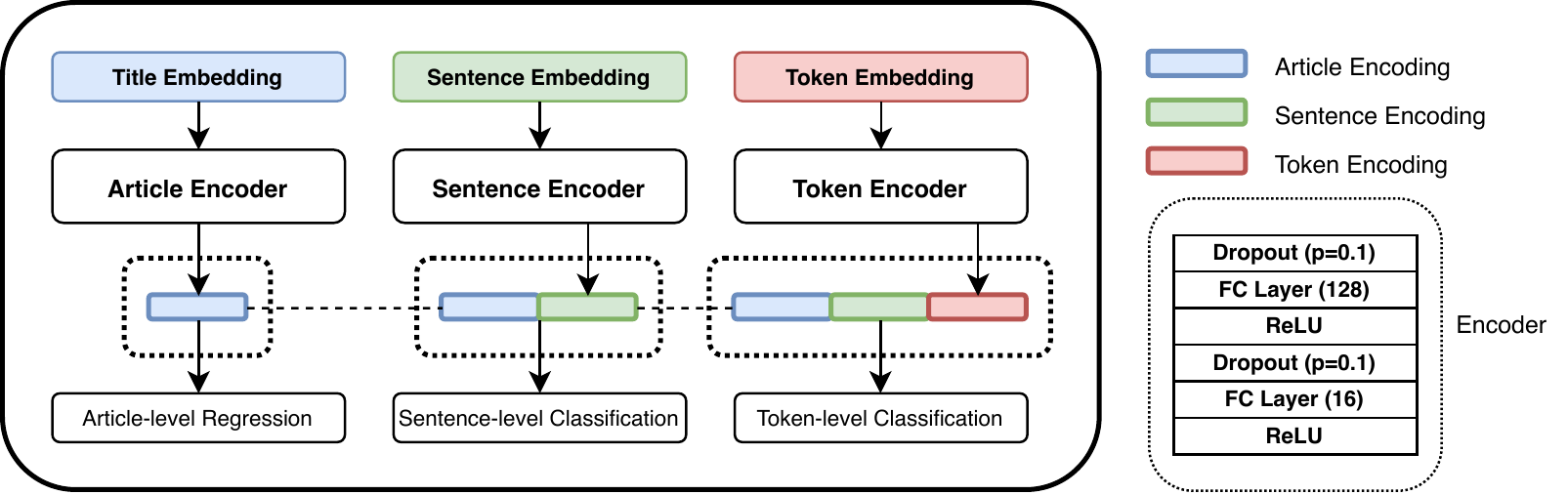}
    \caption{Model Architecture for the Span Identification sub-task.}
    \label{figure1}
\end{figure}

We approach the span identification sub-task with token-level binary classification. We label every token as \emph{belongs / does not belong} to a propaganda span. A continuous sequence of propaganda tokens in the text combine to form a propaganda span. We use pre-trained word embeddings to represent the tokens. As the context from lower granularity is essential during the classification, using only token-level information for this task is not sufficient. To incorporate the lower granularity context for token-classification, we propose a multi-granularity knowledge-sharing model. We use two lower granularity contexts for this: the sentence and the article (document) to which the token belongs. Figure~\ref{figure1} illustrates the high-level architecture of the proposed model for span identification. We train the proposed model simultaneously for three different objectives (one for each granularity): article-level regression, sentence-level classification, and token-level classification. 

\subsubsection{Article-level Regression}
For each article in the dataset, we have the article-title and the article-content split at sentence-level. The title of an article is a good summarized representation of the article's content. We take the mean-pooled embedding of all the word embeddings from the article's title to get the title embedding ($Emb_A$). We represent each article by its respective $Emb_A$. Further, $Emb_A$ is passed through an article encoder (Figure~\ref{figure1}) to get a lower-dimensional article encoding ($Enc_A$). For every article, we calculate the normalized count of sentences with propaganda spans. We refer to this value as the propaganda-ratio ($Ratio_P$) which is defined according to Equation~\ref{equation1}. $Ratio_P$ is restricted to the range $[0,1]$. $Enc_A$ is used for regression over the $Ratio_P$ values of the articles. To minimize the error in predicted $Ratio_P$ values, we use Smooth L1 Loss ($SL1_{A}$) as an objective function for this regression task to avoid exploding gradients from the outlying predictions.

\begin{equation}
    Ratio_P = \frac{Number\ of\ sentences\ with\ propaganda\ spans}{Total\ number\ of\ sentences\ in\ the\ article}
\label{equation1}
\end{equation}

\subsubsection{Sentence-level Classification}
We label every sentence in the article as \emph{contains / does not contain} a propaganda span. For every sentence, we obtain its sentence embedding ($Emb_S$) by taking the mean-pooled embedding of all the word embeddings from the sentence. $Emb_S$ is passed through a sentence encoder (Figure~\ref{figure1}) to get a lower dimensional sentence encoding ($Enc_S$). We concatenate $Enc_S$ with its corresponding $Enc_A$ to incorporate the article context. The concatenated encoding is further used for the sentence-level binary classification task. We use Binary Cross-Entropy Loss ($BCE_{S}$) as an objective function for this task.

\subsubsection{Token-level Classification}
For token-level binary classification, we pass the token word embedding ($Emb_T$) through a token encoder (Figure~\ref{figure1}) to get a lower-dimensional token encoding ($Enc_T$). We concatenate $Enc_T$ with its corresponding $Enc_S$ and $Enc_A$. This concatenated encoding is finally used for the token-level binary classification task. We use Binary Cross-Entropy Loss ($BCE_{T}$) as an objective for this task. The concatenation of encoding with its corresponding lower granularity encoding incorporates a unidirectional knowledge transfer from \emph{lower to higher granularity} level. To implement an implicit bi-directional knowledge transfer, we perform simultaneous training for tasks across all the granularities. To perform the training, we define a combined multi-granularity objective function ($Objective_{MG}$) as shown in Equation~\ref{equation2}. We establish a trade-off between the Smooth L1 Loss ($SL1_{A}$) and the classification losses ($BCE_{S}$ and $BCE_{T}$)  with $\alpha$ as the trade-off factor. Model parameters across all three granularities are trained together to optimize $Objective_{MG}$ objective function, resulting in a multi-task learning setting (Figure~\ref{figure1}).

\begin{equation}
    Objective_{MG} = (\alpha * SL1_{A}) + (1-\alpha)*(BCE_{S} + BCE_{T})
\label{equation2}
\end{equation}

\subsection{Sub-Task 2: Technique Classification}
This sub-task is a multi-class classification of propaganda techniques. Our system classifies each propagandist statement amongst the $14$ given techniques. The dataset is highly imbalanced, and there is a lot of variation in the length of sequences. Therefore, we ensemble our machine learning algorithm and deep learning transformer architecture to provide a solution with better generalization across the classes.

\subsubsection{Machine Learning}
For hand-crafted feature-based machine learning methods, we preprocess the sentences and extract their pragmatic and lexical features. Machine learning algorithms such as XGBoost and Logistic Regression use these features and perform better than the given  baseline substantially and generalize well. Since news articles are written in a formal language with very few typographical or linguistic errors, identifying these features is less technically challenging compared to free-form text. 

We perform preprocessing on our text to remove any unwanted content and enrich our text for feature extraction. The various procedures for the preprocessing include UTF-8 conversion, removal of non-ASCII characters, lower casing, lemmatizing/stemming, and removal of extra white-spaces, newlines, and stop-words. We create the feature space for the classification using three metadata types: contextual (sequence length, count of '!' and '?'), content (word and char n-gram TFIDF, part of speech) and context-based metadata (polarity of a sentence\footnote{TextBlob API. \url{https://textblob.readthedocs.io/en/dev/}}, and topic modeling). However, these models do not effectively capture the sequential nature of the data.

\subsubsection{Deep Learning : Stacked LSTMs, Pretrained Embeddings, CNN-LSTMs}
We shift our focus to deep learning models that retain the sequential information by training Long-Short-Term-Memory [LSTM] Networks \cite{gers1999learning}. The non-linear decision boundaries of these neural networks capture the complex features inside text. We convert the sentences to lowercase, non-ASCII characters are removed and tokenized to words. While validating or testing all obscure words are converted to a special [UNK] (Unknown) token. Approximately $64\%$ of the words present in the development set's vocabulary intersect with the training set's vocabulary. Also, $85.8\%$ of the words in the training-vocabulary have a count lesser than five.

Each word is encoded to fixed-size-vector or embedding. These embeddings are randomly initialized and passed to a stacked LSTM network with a classification layer on top. These networks give us comparable results. For this approach, we use the stacking of two LSTM layers with $256$ hidden-sizes each. On increasing the number of stacked layers and the hidden-size of each LSTM unit, the models tend to overfit and neglect the minority classes due to the relatively small feature space and training samples per class. Therefore to improve the semantic relationship between the words and increase the training vocabulary, we propose systems that have prior knowledge of the language and a larger vocabulary set. Such systems show improvement on these linguistically rich corpora over models with randomly initialized embeddings \cite{erhan2010does}. We can infer this because their language modelling is more refined and wider.

To increase the vocabulary for our model and to extract the contextual importance of words, we incorporate pre-trained word embeddings such as "Global Vectors" [GloVe] \cite{pennington2014glove}. These embeddings map each input token to a fixed-length vector. We set them to be non-trainable to avoid changing these vectors substantially. Pre-trained embeddings help the model to understand the underlying relationships.

Since GloVe is trained on a large corpus to obtain vector representation of words, $98.8\%$ of the vocabulary in the development set and $97.9\%$ of the training vocabulary intersects with its own. Therefore, in total less than $2\%$ of these words are out of vocabulary and address the issue of vocabulary coverage in above models. On inspection, we find that most of the out-of-vocab words in the GloVe are conjunct words commonly used in the text, such as \emph{jawboned, antichrists, trumpists, cyberspies, atleast}. Therefore we use sub-word tokenization and break each of these absent words into two in-vocab words where the smaller word is at least two characters long like \emph{jaw boned, anti christs, trump ists, cyber spies, at least}. The sub-word tokenization increases our coverage to $99.1\%$ with the remaining words either being proper-nouns or non-alphabetic patterns. Since these words convey a little meaning, and are removed during training and testing.

We employ CNNs (Convolutional Neural Networks) and CNN-LSTM models as well by adding convolution filters before extracting and pooling information from the embeddings. CNN-LSTM models help to extract a sequence of higher-level phrase representations \cite{zhou2015c}. As a result of reduced inputs for the LSTM layers, these models converge faster (in less number of epochs) and quicker (in terms of training time per epoch), but the performance does not improve. 

We pad all the sequences which have less than $100$ tokens and truncate the others to the maximum length of $100$. The tokens are encoded to word embeddings. The obtained embeddings are fed to the stacked LSTM layers, with dropout regularization to prevent over-fitting. We pass the LSTM layers' outputs to the linear classification layers with sigmoid activation to obtain the categorical cross-entropy loss. We use Adam optimizer with a small learning rate to update the parameters for $50$ epochs \cite{kingma2014adam}.

\subsubsection{Deep Learning : Transformer Architectures}
Consequently, we transit to deep pre-trained transformer architectures with attention mechanism \cite{vaswani2017attention}. We employ BERT-base \cite{devlin-etal-2019-bert} model which has $12$ Layers. BERT tokenizer implements word-piece tokenization, which eliminates out-of-vocabulary words by splitting the words into sub-words. For example the word "\emph{judgmental}" is broken to ["judgement", "\#\#al"]. This model has been pre-trained on large corpus for masked language modelling and hence can be used for many downstream NLP tasks such as classification, question answering systems, and named-entity-recognition. We freeze most of the layers of the model and train the remaining (layer 10, 11, 12) with a very small learning-rate: $1e$-$4$ for four epochs with a classification layer on top of the extracted pooled output embeddings.

\begin{figure}[htbp]
 \center
 \includegraphics[scale=0.9]{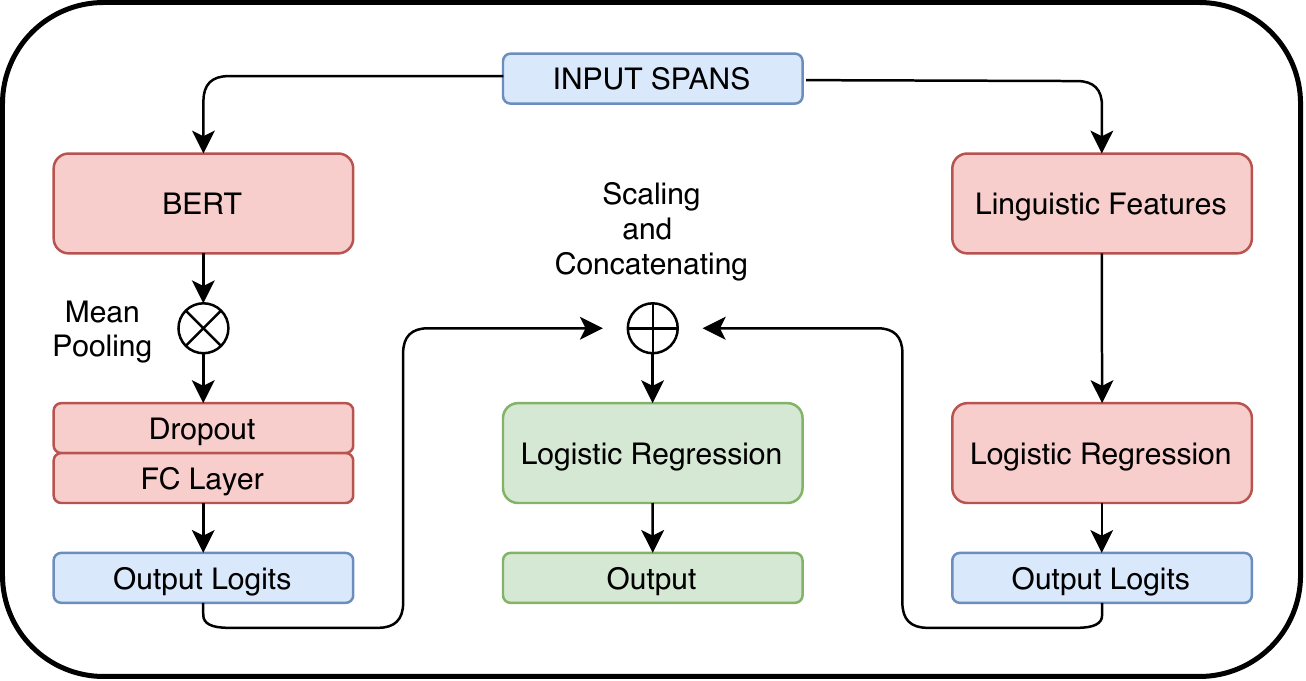}
 \caption{Ensemble Model Architecture for Technique Classification sub-task.}
 \label{figure3}
\end{figure}

\subsubsection{Ensemble Learning}
The majority of the deep learning models perform better than the machine learning models on the validation split. On closer inspection, we observed that deep learning models tended to neglect the minority classes and had some unpromising class-wise F1 scores (discussed in Section~\ref{subsec:result_technique}). On the other hand, machine learning models had no zero-scores for any technique but proved to be less decisive for the majority-classes. We infer that training an ensemble of these models; the proposed architecture can improve in terms of generalization across all the classes. In this manner, we account for the gain in overall performance and also avoid zero-scores for the minority classes. Therefore we concatenate the scaled outputs from both the models and pass this to a logistic regression classifier. We penalize this classifier by the L2-norm (euclidean distance) and keep class wise weights for the objective in inverse proportion to their count in training set to obtain the final predictions. Our F1 scores for minority classes show massive improvements while only a minor change $(+0.006)$ in the overall micro-F1 score.

\section{Experimental Setup}\label{sec:experimental_setup}
\subsection{Sub-Task 1: Span Identification}
For the span-identification sub-task, we use word embeddings from pre-trained transformer models \cite{vaswani2017attention}, extracted using \textit{flairNLP}\footnote{Flair. A state-of-the-art NLP framework. \url{https://github.com/flairNLP/flair}} \cite{akbik2018coling}. We test our model with various embeddings such as BERT \cite{devlin-etal-2019-bert}, RoBERTa \cite{DBLP:journals/corr/abs-1907-11692}, XLNet \cite{DBLP:journals/corr/abs-1906-08237} and GPT-2 \cite{radford2019language}, which differ in their model architectures and pre-training methods. These embeddings and their respective transformer-models are kept non-trainable throughout our experiments so that the standalone performance of our proposed system can be monitored effectively. We split the given training dataset at token-level with a $95$-$5$ split for training and validation purposes, respectively. We compute the accuracy, F1 score, precision, and recall metrics for the token-level binary-classification. Since there is a high class-imbalance for token-level classification, we consider the token-level F1 score for validation purposes. The official evaluation metrics (span-level F1, precision, and recall) used for this task are calculated with respect to the overlaps between the predicted and the ground-truth spans \cite{da-san-martino-etal-2019-fine}.

\subsection{Sub-Task 2: Technique Classification}
For this sub-task, we use BERT (base-uncased variant), a pre-trained transformer model based on the work done by \newcite{devlin2018bert} with a classification layer on top. We extract the contextual pooled output embedding from the BERT model using \textit{transformers} \footnote{Transformers. \url{https://github.com/huggingface/transformers}}, a python package by Hugging Face. All layers except for the last two and classification layers are kept non-trainable. To avoid updating the saved state of the parameters substantially, we train it with a small learning rate for four epochs, it increases in $1000$ steps to $1e-4$ and then decays linearly. The gradients are also clipped as discussed by \newcite{DBLP:journals/corr/abs-1904-00962} and the AdamW optimizer is used to optimize the model \cite{DBLP:journals/corr/abs-1711-05101}. For machine learning models, we examine the feature importance scores from the XGBoost model for feature selection \cite{zheng2017short}. We only consider features that have relative feature importance scores more than $0.30$, i.e. word $n$-grams \emph{tf-idf} scores for $n$ = $\{1, 2, 3\}$ and character $k$-gram \emph{tf-idf} scores for $k$ = $1$ to $6$, and the character count of the spans. We use the logistic regression model with L2 penalty and class weights inversely proportional to their count in the training set to get the technique classification probabilities. The output logits from BERT's classification layer are scaled (min-max scaling) and concatenated with the scaled probabilities obtained from the logistic regression model. We pass this concatenated vector to another logistic regression classifier with the same hyper-parameters to get the final predictions.

\section{Results}

\begin{table}[htbp]
\centering
\resizebox{0.8\textwidth}{!}{%
    \begin{tabular}{@{}cccccccc@{}}
        \toprule
        \multirow{2}{*}{\textbf{\begin{tabular}[c]{@{}c@{}}Embedding\end{tabular}}} & \multicolumn{3}{c}{\textbf{Span-level Metrics}} & \multicolumn{4}{c}{\textbf{Token-level Metrics}} \\ \cmidrule(l){2-8} 
         & \textbf{Precision} & \textbf{Recall} & \textbf{F1} & \textbf{Precision} & \textbf{Recall} & \textbf{F1} & \textbf{Accuracy} \\ \midrule
        BERT-Token & 0.319 & 0.374 & 0.344 & 0.601 & 0.333 & 0.428 & 0.885 \\ \midrule
        RoBERTa-Token & 0.358 & 0.375 & 0.366 & 0.62 & 0.233 & 0.339 & 0.883 \\ \midrule
        BERT-MG & 0.304 & 0.409 & 0.349 & 0.846 & 0.786 & \textbf{0.815} & \textbf{0.952} \\ \midrule
        RoBERTa-MG\textsuperscript{*} & \textbf{0.347} & \textbf{0.391} & \textbf{0.368} & \textbf{0.852} & 0.779 & 0.814 & \textbf{0.952} \\ \midrule
        XLNet-MG & 0.304 & 0.413 & 0.35 & 0.82 & \textbf{0.794} & 0.807 & 0.949 \\ \midrule
        GPT-2-MG & 0.289 & 0.406 & 0.337 & 0.805 & 0.752 & 0.778 & 0.942 \\ \bottomrule
    \end{tabular}%
}
\caption{Results for the Span Identification sub-task on the development set (*Final submission)}
\label{task-1-results}
\end{table}

\subsection{Sub-Task 1: Span Identification}
For the span-identification sub-task, we consider token-only baselines with \emph{BERT-Token} and \emph{RoBERTa-Token} models. These models only perform token-level binary-classification task and do not consider the sentence-level and article-level granularities. We compare these baselines to our proposed multi-granularity model (\emph{BERT-MG}, \emph{RoBERTa-MG}, \emph{XLNet-MG} and \emph{GPT-2-MG}). Table~\ref{task-1-results} shows the Span-level metrics for the official development set and the token-level metrics on our token-level validation split. Our multi-granularity model consistently achieves more than $94\%$ token-level accuracy across all the four types of embeddings as compared to the $88\%$ token-level accuracy shown by the token-only baselines. The multi-granularity model also shows some improvement on the span-level F1 score as compared to the token-only baselines. We get the best token-level F1 scores $(0.815)$ with BERT and RoBERTa embeddings. RoBERTa embeddings outperforms all the other embeddings for the span-level metrics. Therefore, we use RoBERTa embeddings for the final submission to the task leaderboard. The performance for this sub-task can possibly be improved further by using trainable embeddings by fine-tuning their respective transformer-models. Our system ranks $21^{st}$ on the final test-set leaderboard with a span-level F1 score of $0.387$.

\begin{table}[htbp]
\centering
\resizebox{0.7\textwidth}{!}{%
    \begin{tabular}{@{}cccccccc@{}}
        \toprule
        \multirow{2}{*}{\textbf{\begin{tabular}[c]{@{}c@{}}Model\end{tabular}}} & \multicolumn{2}{c}{\textbf{Validation Split}} & \multicolumn{2}{c}{\textbf{Development Set}} \\ \cmidrule(l){2-5} 
         & \textbf{Micro F1} & \textbf{Macro F1} & \textbf{Micro F1} & \textbf{Macro F1} \\ \midrule
        Logistic Regression & 0.54 & 0.42 & 0.51 & 0.32 \\ \midrule
        LSTM & 0.55 & 0.32 & 0.49 & 0.29 \\ \midrule
        LSTM (Glove) & 0.59 & 0.36 & 0.55 & 0.36 \\ \midrule
        CNN-LSTM (Glove) & 0.56 & 0.28 & 0.48 & 0.26 \\ \midrule
        BERT & \textbf{0.66} & 0.42 & \textbf{0.58} & 0.38 \\ \midrule
        BERT + Logistic Regression\textsuperscript{*} & 0.65 & \textbf{0.45} & \textbf{0.58} & \textbf{0.43} \\ \midrule
        RoBERTa & 0.63 & 0.38 & 0.54 & 0.32 \\ \bottomrule
    \end{tabular}%
}
\caption{Results for the Technique Classification sub-task on the development set and validation split (*Final submission)}
\label{task-2-results}
\end{table}
\begin{table}[h]
\centering
\resizebox{0.9\textwidth}{!}{%
    \begin{tabular}{@{}lccccccc@{}}
        \toprule
         & \multirow{2}{*}{\textbf{LR}} & \multirow{2}{*}{\textbf{LSTM}} & \multirow{2}{*}{\textbf{\begin{tabular}[c]{@{}c@{}}LSTM \\ (Glove)\end{tabular}}} & \multirow{2}{*}{\textbf{\begin{tabular}[c]{@{}c@{}}CNN-LSTM\\ (Glove)\end{tabular}}} & \multirow{2}{*}{\textbf{BERT}} & \multirow{2}{*}{\textbf{BERT-LR\textsuperscript{*}}} & \multirow{2}{*}{\textbf{RoBERTa}} \\ 
         & & & & & & & \\ \midrule
        Appeal\_to\_Authority & 0.063 & 0.077 & 0.000 & 0.000 & 0.000 & 0.400 & 0.118 \\ \midrule
        Appeal\_to\_fear-prejudice & 0.160 & 0.205 & 0.292 & 0.321 & 0.376 & 0.372 & 0.317 \\ \midrule
        Bandwagon, Reductio\_ad\_hitlerum & 0.286 & 0.000 & 0.431 & 0.000 & 0.500 & 0.484 & 0.000 \\ \midrule
        Black-and-White\_Fallacy & 0.167 & 0.051 & 0.154 & 0.000 & 0.154 & 0.145 & 0.000 \\ \midrule
        Causal\_Oversimplification & 0.341 & 0.341 & 0.358 & 0.143 & 0.250 & 0.258 & 0.320 \\ \midrule
        Doubt & 0.444 & 0.431 & 0.493 & 0.497 & 0.517 & 0.479 & 0.503 \\ \midrule
        Exaggeration, Minimisation & 0.371 & 0.313 & 0.443 & 0.417 & 0.418 & 0.377 & 0.447 \\ \midrule
        Flag-Waving & 0.679 & 0.671 & 0.714 & 0.663 & 0.728 & 0.735 & 0.745 \\ \midrule
        Loaded\_Language & 0.712 & 0.662 & 0.703 & 0.691 & 0.725 & 0.720 & 0.699 \\ \midrule
        Name\_Calling, labelling & 0.591 & 0.539 & 0.645 & 0.682 & 0.693 & 0.700 & 0.691 \\ \midrule
        Repetition & 0.350 & 0.198 & 0.251 & 0.099 & 0.264 & 0.412 & 0.0546 \\ \midrule
        Slogans & 0.190 & 0.400 & 0.485 & 0.070 & 0.523 & 0.501 & 0.419 \\ \midrule
        Thought-terminating\_Cliches & 0.061 & 0.121 & 0.000 & 0.000 & 0.167 & 0.240 & 0.111 \\ \midrule
        Whataboutism, Straw\_Men, Red\_Herring & 0.100 & 0.053 & 0.059 & 0.067 & 0.049 & 0.292 & 0.0588 \\ \midrule \textbf{micro F1} & 0.514 & 0.490 & 0.554 & 0.485 & 0.578 & \textbf{0.584} & 0.550 \\ \midrule
        \textbf{macro F1} & 0.322 & 0.290 & 0.359 & 0.260 & 0.383 & \textbf{0.436} & 0.320 \\ \bottomrule
    \end{tabular}%
}
\caption{Class-wise F1 scores on the development set (*Final submission)}
\label{class-wise}
\end{table}
\subsection{Sub-Task 2: Technique Classification}\label{subsec:result_technique}
For the technique-classification sub-task, our system ranks $18^{th}$ with a micro-F1 score of $0.54$ on the final test-set. We have shown the performances of our models on the development set in Table~\ref{task-2-results}. Since this is an imbalanced multi-class classification task, we also report our models' macro F1 scores to test our proposed systems' generalization across all the propaganda techniques. Macro averaging gives equal importance to all classes. Table~\ref{task-2-results} unveils that in both validation and development splits, the combined BERT and logistic regression outperforms all models with a macro-f1 score of $0.45$ and $0.43$, respectively.

To further inspect the performance of our models across particular techniques, we also calculate the class-wise F1 scores as shown in Table~\ref{class-wise}. Machine learning models (logistic regression) achieve a relatively lower micro averaged F1 score of $0.514$ on the development set, but give non-zero scores for all the classes. Stacked LSTM networks with randomly initialized embeddings do not achieve any improvement $(0.490)$ on the development-set and conceive near-zero scores for multiple classes as well. This is due to the vast vocabulary of this relatively smaller dataset. Consequently, on using pre-trained GloVe embeddings with the stacked LSTM networks, we achieve a significant improvement $(0.52)$.

Similarly, while CNN-LSTM models are relatively faster, they lower the overall performance $(0.485)$ and conceive zero f1 scores for multiple classes. Transformer based models, namely BERT $(0.578)$ and RoBERTa $(0.550)$ outperform the above models significantly; however, they do not perform well for the minority classes. In fact, BERT gives zero or near zeros scores for some labels like  \textit{Appeal\_to\_Authority; Whataboutism, Straw\_Men, Red\_Herring}.

The ensemble of BERT with the logistic regression model does not improve the micro F1 score significantly $(0.584) $(1\% gain) but unlike other models it gives non-zero F1 score for all the classes, improving the macro F1 score from $0.38$ to $0.43$ ($13.16\%$ gain).
The F1 scores increase from $0.0$ to $0.4$ for \textit{Appeal\_to\_Authority} and from $0.049$ to $0.292$ for \textit{Whataboutism, Straw\_Men, Red\_Herring}.
We use this ensemble model as our final submission due to its better generalization across all the propaganda technique classes. Our results reveal that minority classes such as \textit{Appeal to authority, Bandwagon, Reductio ad Hitlerum, Black and white fallacy} and \textit{Thought terminating cliches} are relatively difficult to predict. 

\section{Reproducibility}
We make our code publicly available as GitHub repositories. The source code for the proposed multi-granularity knowledge sharing model for sub-task $1$ can be accessed at (\url{https://github.com/rajaswa/semeval2020-task11}). The source code for the proposed BERT ensemble with linguistic features based logistic regression model for sub-task $2$ can be accessed at (\url{https://github.com/someshsingh22/News-Propaganda-Detection}). Our models and acquired results are available for benchmarking, comparison, and reproducibility. We do not share the experimental dataset as per the task guidelines.

\section{Conclusion}
In this paper, we proposed systems for the task of span identification and multi-class imbalanced technique classification of propaganda spans in news articles. We analyzed the performance of various machine learning and deep learning-based architectures for these high granularity tasks. On the span-identification sub-task test set, our multi-granularity knowledge sharing model gives a span-level F1 score of $0.387$. For the technique classification task, our ensemble of pre-trained transformer model with logistic regression gives a micro F1 score of $0.54$. We further infer the effectiveness of incorporating linguistic features and achieve non-zero F1 scores for all techniques and $13.6\%$ gain in the macro-F1 score. Our results also unveil the limitations/ineffectiveness of deep learning models to capture the minority-class techniques. 

We plan to extend our work by using trainable transformer-model embeddings and improve the performance of the span-identification sub-task. The work can further be enhanced by adding more granularities for knowledge-sharing. The proposed knowledge-sharing model may also be used for various closely-related tasks such as fake news and hate speech detection, given application-specific appropriate objective functions are defined across multiple granularities for these tasks.

\section*{Acknowledgements}
We would like to thank the reviewers for their timely and constructive feedback on our work and the shared-task organizers for the opportunity and task-related resources. We would also like to thank the SemEval-2020 workshop organizers for the timely follow-ups and the workshop organization.

\bibliographystyle{coling}
\bibliography{semeval2020}
\end{document}